\documentclass[10pt,twocolumn,letterpaper]{article}

\usepackage{iccv}              

\definecolor{iccvblue}{rgb}{0.21,0.49,0.74}
\usepackage[pagebackref,breaklinks,colorlinks,allcolors=iccvblue]{hyperref}

\def\paperID{*****} 
\def\confName{ICCV}
\def\confYear{2025}

\title{Federated Foundation Models Raise New Concerns of \\ Robustness, Privacy, and Fairness}

\author{Jiaqi Wang\thanks{Equal contribution. This paper has been accepted by TrustFM: Workshop on Trustworthy Foundation Models
in conjunction with ICCV 2025.} \\
Auburn University\\
{\tt\small jqwang@auburn.edu}
\and
Xi Li\footnotemark[1]\\
University of Alabama at Birmingham\\
{\tt\small xiliuab@uab.edu}
}

\begin{document}
\maketitle

\begin{abstract}
Incorporating foundation models (FMs) into federated learning (FL) offers a mutual benefit for foundation models and local clients. However, this incorporation introduces novel issues in terms of robustness, privacy, and fairness, which have not been sufficiently addressed in the existing research. We make a preliminary investigation into this field by systematically evaluating the implications of FM-FL integration across these dimensions. We argue that robustness, privacy, and fairness must be rethought and re-ensured in the context of FM–FL systems.
\end{abstract}
\vspace{-4mm}
\vspace{-0.05in}
\section{Introduction}

Federated Learning (FL) has emerged as a vital framework for collaborative model training across multiple devices while preserving data privacy. It is widely applied in domains such as healthcare\cite{WangQCGM22}, education\cite{fachola2023federated,mistry2023privacy}, and finance\cite{imteaj2022leveraging,shi2023responsible}, enabling local model training and sharing only updates to enhance privacy and efficiency. Despite its advantages, FL faces challenges such as ineffective training and resource constraints. 
Ineffective training in FL arises from limited local datasets, requiring multiple rounds of aggregation to achieve a global model comparable to centralized training \cite{imteaj2021survey,bouacida2021vulnerabilities}.
This issue is more pronounced in personalized FL, where heterogeneous data distributions hinder model convergence. 
Additionally, the distributed nature of FL introduces resource disparities among clients, impacting training efficiency and scalability \cite{shi2020joint,wang2023knowledge,khan2021federated,zhang2022federated}. Devices with lower capabilities may contribute less effectively, and simpler models may offer limited improvements to the global model \cite{liu2022contribution,zeng2021fedcav}.

\begin{figure}[t]
    \centering
    \begin{minipage}{0.23\textwidth}
        \centering
        \vspace{-0.1in}
        \includegraphics[width=.85\textwidth]{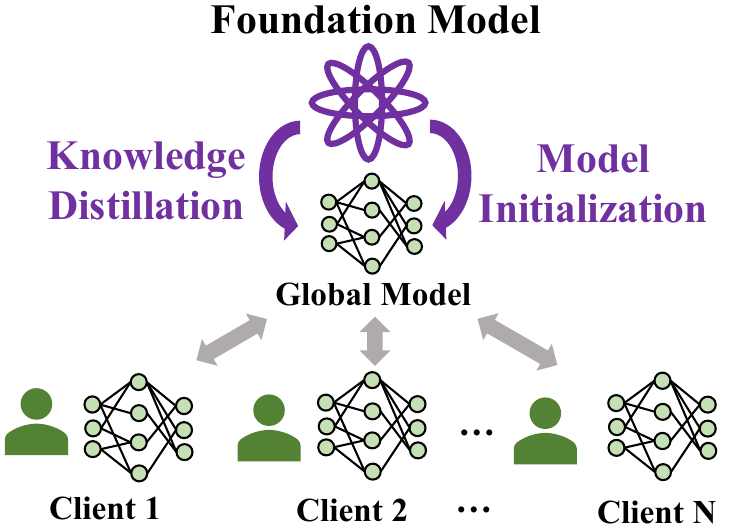}
        \caption{Foundation models at the server side.}
        \label{fig:server}
        \vspace{-0.1in}
    \end{minipage}
    \hfill
    \begin{minipage}{0.24\textwidth}
        \centering
        \includegraphics[width=.90\textwidth]{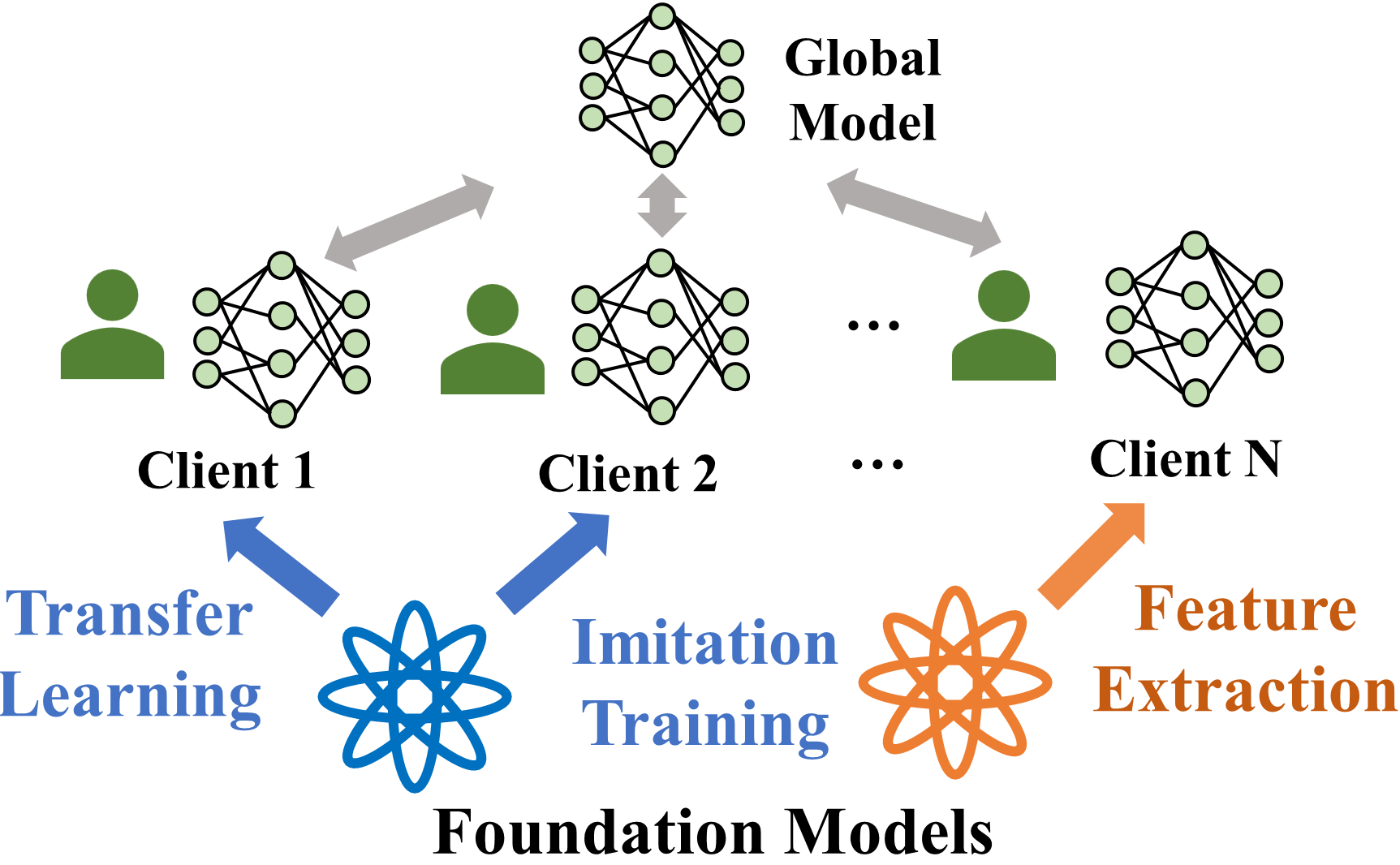}
        \caption{Foundation models at the client side.}
        \label{fig:client}
        
    \end{minipage}
    \vspace{-0.2in}
\end{figure}

Foundation Models (FMs), known for their vast knowledge and versatility, offer a promising solution to these challenges. 
FMs can mitigate issues of ineffective training by, \textit{e.g.}, producing data that mirrors real-world distributions, to improve FL models' performance and generalization \cite{eigenschink2021deep}. 
FMs can also significantly enhance the performance of FL systems by serving as teachers \cite{XingXQZZ22,ZhuHZ21,FedMD,abs-2303-10917} to transfer FMs' extensive and diverse understanding. 
Furthermore, FMs facilitate a reduction in computational burden through transfer learning, allowing clients to fine-tune pre-trained models with their local data, thereby requiring less computational power \cite{zhuang2023foundation}. 
Additionally, FMs can minimize communication overhead by serving as efficient encoders, reducing the need for extensive data transmission during the model aggregation phase\cite{yi2023fedgh,kalra2023decentralized}.
The integration of FMs in to FL (FM-FL) attracts attention in multiple domains and the integration significantly enhances both efficiency and effectiveness\cite{zhuang2023foundation}. Overall, the FMs could be integrated on the server~\cite{zhang2023gpt} or deployed at the clients~\cite{lu2023fedclip,chen2024feddat}.
As shown in Fig.~\ref{fig:server}, the FM is integrated on the server for tasks such as model pre-training and knowledge distillation\cite{FedDF, FedMD}.
The FMs can also be deployed at clients, as shown in Fig.~\ref{fig:client}. 

Despite the potential of FM-FL in various domains, ensuring that AI models in FM-FL are responsible and reliable is both important and challenging. 
There is a noticeable research gap in exploring this issue. 
The interaction between FMs and FL introduces new dimensions of challenges concerning robustness, privacy, and fairness. 
These concerns differ from existing threats against FL and can evade current countermeasures. We argue that robustness, privacy, and fairness must be rethought and re-ensured in the context of FM–FL systems. This calls for a paradigm shift: future research should move beyond adapting existing FL or FM techniques and instead develop holistic, interaction-aware approaches to model and mitigate emergent threats. Our contribution is summarized as the following: 
\begin{enumerate}[leftmargin=5mm]
    \item Building on existing FM-FL frameworks, we identify novel dimensions introduced by integrating FMs into FL, focusing on robustness, privacy, and fairness. 
    \item We conduct comprehensive investigations into these novel threats, analyzing how they differ from traditional FL, evaluating the potential failure of current countermeasures, and identifying promising research directions to assess and address these emerging challenges.
    \item Lastly, we provide an extensive discussion on the responsibility implications of FM-FL, exploring how this integration offers new perspectives on addressing problems, advances our state of knowledge, and impacts the broader field of machine learning, not just FL.
\end{enumerate}

\begin{figure*}[t!]
\centering
\includegraphics[width=0.88\textwidth]{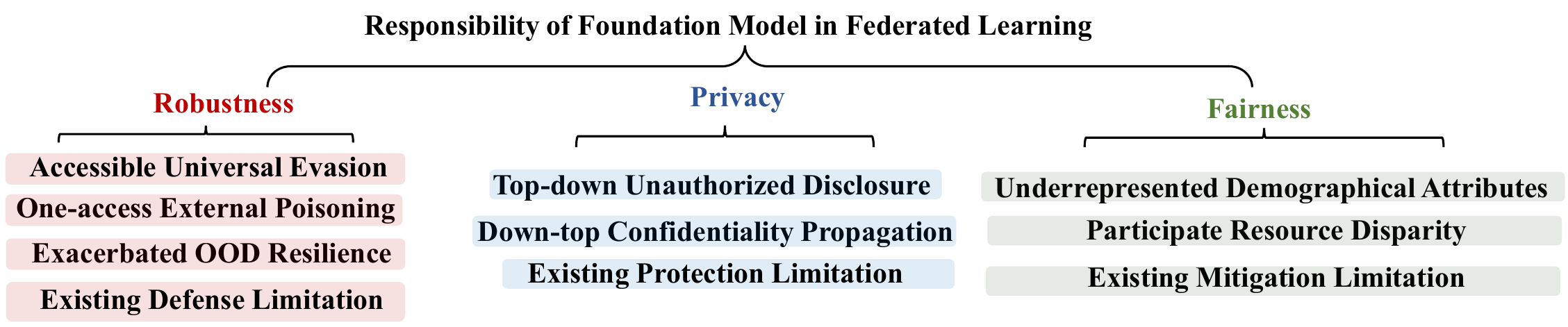}
    \caption{Overview of the FM-FL responsibility: robustness, privacy, and fairness.}
    \label{fig:overview}
    \vspace{-0.2in}
\end{figure*}


\section{Robustness}\label{sec:robustness}
In this section, we explore new mechanisms of robustness challenges in FM-FL, focusing on 
accessible universal evasion, one-access external poisoning, and exacerbated OOD resilience.

\noindent\textbf{Accessible Universal Evasion}
The proposed evasion attack assumes the attacker compromises at least one client and exploits black-box access to the FM, crafting intelligent, universal ``jailbreak'' prompts without requiring optimization knowledge. Unlike classic, domain-specific adversarial attacks with limited transferability, these natural language prompts generalize across modalities and aim to undermine the entire FL system. In LLM-FL settings, where clients use prompts to generate synthetic data for privacy-preserving training, compromised clients can inject harmful prompts that cause the LLM to produce biased or misleading content. Once shared, this corrupted data degrades model robustness and jeopardizes system integrity.

\noindent\textbf{One-Access External Poisoning} Unlike traditional poisoning attacks in FL that require compromising clients and maintaining persistent participation, this novel threat allows an external third party -- such as an LLM service provider -- to compromise the integrated FM without direct access to clients or the server. With white-box access, the attacker can fine-tune the FM using poisoning instructions \cite{BD_instruction_LLM}, or with black-box access, insert malicious system prompts \cite{DecodingTrust}. In practical scenarios where LLMs are used for synthetic data generation and knowledge transfer, third-party APIs are often leveraged for efficiency and scalability \cite{api_integration,cost_reduction,prompt_engineering}. A malicious provider can inject harmful instructions that taint the FM, leading to poisoned content during initialization and interaction phases, which then propagates through model aggregation \cite{FedDF,FedMD,BD_FMFL,BD_FMHFL}. This process ultimately undermines the integrity and robustness of the entire FL system.

\noindent\textbf{Exacerbated Out of Distribution Resilience} Integrating FMs into FL can intensify OOD challenges, especially when FL tasks fall outside the FM’s domain expertise. In synthetic data generation, limited access to diverse client data prevents effective FM adaptation, leading to synthetic samples that misrepresent client distributions or contain nonsensical content, which degrades training performance and slows convergence. Similarly, in knowledge distillation, if the FM lacks relevant knowledge, it may produce inaccurate or fabricated information, resulting in flawed supervision that harms FL model performance.




\noindent\textbf{Limitation of Existing Defense 
} Integrating FMs into FL introduces new threats that existing defenses are ill-equipped to handle. Accessible universal evasion attacks exploit LLM vulnerabilities using jailbreak prompts during training, bypassing the need for input-specific perturbations and undermining traditional adversarial defenses. One-access external poisoning embeds threats into FMs through a single compromise, enabling wide-scale impact without persistent attacker involvement. This method avoids detection by using clean local datasets, rendering typical outlier-based defenses ineffective. FM integration also worsens OOD issues, as synthetic data and distilled knowledge from FMs lacking task-specific expertise can misalign with FL data, harming convergence and model performance.




\section{Privacy}\label{sec:privacy}
In this section, we delve into emerging privacy challenges within FM-FL top-down unauthorized disclosure and bottom-up confidentiality propagation. 
\noindent\textbf{Top-down Unauthorized Disclosure} In scenarios where a central FM (at the server) is utilized to enhance or initialize local models within an FL framework, there is a risk of transmitting sensitive information embedded within the FM. This can occur through:
(1) Synthetic Data Generation: Sensitive information might be embedded in synthetic data generated by the FM and transmitted to clients; and
(2) Knowledge Distillation: FM used as a teacher can transfer sensitive information to FL models.
Additionally, FL models initialized with FM parameters and fine-tuned on local datasets may also retain and memorize sensitive information.


\noindent\textbf{Down-top Confidentiality Propagation} In scenarios where FMs are deployed to local clients to assist training, multiple clients might access the same FM due to resource limitations. 
Clients may request synthetic data from the FM, embedding sensitive details reflective of their own datasets. The FM can inadvertently memorize this information, posing a privacy risk. When another client accesses the same FM, there is a risk of unintentionally exposing the sensitive information. This highlights a significant privacy concern, as sharing FMs among clients can lead to unintended information leakage across the federated network.


\noindent\textbf{Existing Protection Limitation} In FL, privacy leakage predominantly occurs from local clients to the central server.
Sensitive information, such as membership, data distribution, and even specific training data, are reverse-engineered from local updates through inference attacks.
However, for FM-FL, the leakage is more complex, involving not just the transfer of model information, but also the potential embedding and transmission of sensitive data within synthetic data generated by FMs or through knowledge distillation.
FM-FL introduces a two-way leakage risk: from FM to FL (where sensitive information within the FM can be transferred to FL clients) and from FL to FM (where sensitive local client data can be memorized by the FM during queries for synthetic data or knowledge).
the advanced capabilities of FMs and the interaction between FM and FL makes it a more dynamic and multifaceted privacy challenge.

\section{Fairness}\label{sec:fairness}

In this section, we explore emerging fairness challenges in FM-FL: underrepresented demographic attributes and participation resource disparity. 


\noindent\textbf{Underrepresented Demographical Attributes} Demographical Attributes, such as gender, race, and age, define diverse groups. 
Underrepresented attributes in datasets lead to biased outputs and inequitable results for marginalized groups. This concern is heightened with FMs, which often learn inherent biases from extensive, internet-sourced datasets.
Integrating FMs into FL can significantly impact underrepresented demographical attributes. The first challenge is transmission of inherent unfairness from FMs to FL. Using FMs for synthetic data generation and knowledge distillation in FL risks transmitting existing biases to FL models. The second challenge is unfairness injected into FM from compromised FL clients. Adversaries can introduce unfair content through prompts, leading to biased model outputs. 

\noindent\textbf{Participate Resource Disparity} It refers to the unequal distribution of resources -- such as data volume, computational power, memory, connectivity, and funding -- among clients in FL, categorizing them into resource-abundant and resource-constrained groups. This imbalance affects clients' ability to process data, train models, and contribute meaningfully to the collaborative process, leading to unequal influence and benefits. The integration of FMs, which require significant computational and memory resources, exacerbates this disparity by amplifying differences in local training capacity and communication frequency. In FM-FL settings, resource-abundant clients gain substantial advantages through local FM access, enabling enhanced training and more active participation. Their contributions often dominate the global model, skewing it toward their data distributions. In contrast, resource-constrained clients, relying on shared APIs with limited funding and access, struggle to utilize FM capabilities, limiting both their contributions and the benefits they receive. This asymmetry perpetuates systemic unfairness and deepens the participation gap in FL.

\noindent\textbf{Existing Mitigation Limitation} Fairness challenges in FM-FL are more complex than in traditional FL due to additional bias sources introduced by foundation models. While FL primarily suffers from data heterogeneity and uneven client participation, FM-FL inherits biases from the FM’s pretraining and in-context learning, which are amplified through synthetic data and knowledge distillation. This can lead to aggregation bias, where resource-rich clients exert disproportionate influence, marginalizing under-resourced clients and undermining participation fairness. Although many aggregation strategies aim to optimize performance by weighting client contributions based on resources \cite{WangQCGM22,Deng0RCYZZ21,NishioY19}, they often overlook fairness, especially in heterogeneous settings, resulting in biased models that offer limited benefit to disadvantaged participants. Addressing these disparities is essential to preserve the collaborative and equitable spirit of FL.

\section{Future Direction}\label{sec:future}

Considering the integration of FMs into FL impacts robustness, privacy, and fairness, future research should focus on evaluating these effects and devising solutions. Interdisciplinary efforts combining machine learning, cybersecurity, and ethics are vital for developing robust, transparent, and ethical FL systems, necessitating ongoing innovation and societal impact assessments.

\subsection{Robustness}

\noindent\textbf{Assessing the Susceptibility of FM-FL under the novel threats}
A comprehensive assessment of FM-FL's susceptibility to novel threats is essential. This evaluation should address the effects of evasion attacks, particularly through ``jailbreak'' prompts in LLMs, on FL model performance, convergence, misinformation spread, societal impacts, and attacker cost-effectiveness. It must also investigate the influence of compromised FMs, such as backdoor attacks, on FL integrity, focusing on vulnerabilities introduced by synthetic data. Assessing the attackers' efforts in poisoning FL systems, including FM fine-tuning and exploiting ICL, is critical. Finally, measuring the discrepancy between FM-generated and actual client data, alongside the FM's performance on FL-specific tasks, is vital for identifying enhancements in data generation and knowledge distillation to boost FL robustness and reliability.

\noindent\textbf{Effectiveness of Current FL Defenses}
A comprehensive evaluation of the current defense mechanisms in FL against emerging threats is essential. This assessment should encompass the effectiveness of robust aggregation strategies and post-training detection methods in countering these novel threats. Such evaluations provide insights into the adequacy of the defense approaches and the necessity for the robust defensive strategies for FM-FL.

\noindent\textbf{Strengthening the Robustness of FM-FL}
Explore prompt validation techniques to identify and eliminate malicious inputs before they are processed by FMs, such as anomaly detection to spot harmful prompts and prevent adversarial data manipulation. Address the limitations of current FL defenses designed for decentralized threats, recognizing that new attacks target FMs directly and do not appear as statistical outliers. Develop defenses against centralized attacks and those compromising numerous clients without relying on anomaly detection. Investigate dynamic knowledge distillation methods tailored to FL's varied tasks, enhancing knowledge transfer where FMs are confident and leveraging learner models for areas where FMs are less certain, ensuring effective learning across FL scenarios.

\vspace{-0.05in}
\subsection{Privacy}

\noindent\textbf{Privacy Exposure Assessment}
This involves assessing privacy leakage by evaluating sensitive data in FM-generated synthetic datasets, transfer mechanisms during knowledge distillation, and privacy risks in FL models fine-tuned with local data. It also includes investigating vulnerabilities to inference attacks to determine if leaked sensitive information can be reverse-engineered.

\noindent\textbf{Privacy Enhancement Techniques}
This involves developing methods to detect and anonymize sensitive information while preserving data utility for FL models. The focus is on balancing privacy preservation and data richness for robust model development. Additionally, it explores designing algorithms to unlearn or discard sensitive data post-training, ensuring adaptive privacy protection without compromising performance.

\noindent\textbf{Privacy-Preserving Knowledge Transfer and Effective Unlearning Algorithms}
Develop adaptive knowledge distillation algorithms to obscure sensitive data details. Design efficient unlearning algorithms to enable FL systems to discard sensitive data during or after training, minimizing the impact on computational resources and overall performance, thereby maintaining operational integrity while safeguarding privacy.

\noindent\textbf{Ethical Guidelines and Policy Development for FM-FL}
Formulate ethical guidelines for FM-FL integration to address privacy challenges. Develop policy frameworks to govern data collection, use, and sharing within FM-FL systems, balancing innovation and privacy.

\vspace{-0.05in}
\subsection{Fairness}
To foster a more equitable and bias-aware FM-FL ecosystem, we propose a series of strategic research directions aimed at addressing fairness issues in the FM-FL: 

\noindent\textbf{Bias Transmission Assessment from FMs to FL}
Thoroughly examine how biases in synthetic data and knowledge distillation affect FL model outputs, particularly for underrepresented groups. Understanding this bias transfer helps create strategies to make FL models fairer and more representative.

\noindent\textbf{Fair Data Creation and Knowledge Distillation}
Develop methods to scan and neutralize biases in FM-produced data, ensuring equitable representation for all demographic groups. Innovate knowledge distillation techniques to identify and correct biases, promoting fair knowledge transfer to FL models. This dual strategy enhances FL system integrity and inclusivity by reducing bias in data generation and training.


\noindent\textbf{Equitable Participation and Collaborative Learning}
This includes understanding how contributions from well-resourced clients might disproportionately shape FL model development and outcomes, potentially skewing performance and fairness to the detriment of resource-limited clients.
Besides, we propose equitable participation mechanisms and collaborative learning models to bridge the gap between resource-diverse clients. Strategies like round-robin participation and fair access policies to FMs ensure all clients contribute to and benefit equally from FL. Additionally, we suggest frameworks allowing resource-constrained clients to leverage the capabilities of resource-abundant counterparts, efficiently transferring knowledge to smaller, communication-efficient models. This dual strategy democratizes FM access in FL, enhancing the robustness, fairness, and performance of FMFL systems.

\noindent\textbf{Formulation of Fairness-Driven Policies in FM-FL}
Policies should mitigate biases and ensure equitable participation in FM-FL, fostering an inclusive digital ecosystem. They must prioritize resource distribution, fair data representation, and fairness metrics evaluation, guiding FM-FL systems towards technological excellence that benefits a diverse society and promotes a fair, accessible future for all.

\vspace{-0.05in}
\section{Conclusion} \label{sec:conclusion}
In conclusion, this paper presents an in-depth analysis of the robustness, privacy, and fairness challenges arising from the integration of FMs with FL, which remains underexplored in current research. We shed lights on potential future directions in this field, advocating for the development of a responsible FM-FL ecosystem. This ecosystem would not only leverage the strengths of FMs to enhance FL but also prioritize the establishment of secure, reliable, and ethical practices to safeguard against any adverse implications of their deployment. 


{
    \small
    \bibliographystyle{ieeenat_fullname}
    \bibliography{main}
}

\end{document}